\documentclass{article}

\usepackage{spconf}
\usepackage{amsmath,amssymb,amsfonts}
\usepackage{algorithmic}
\usepackage{graphicx}
\usepackage{textcomp}
\usepackage{wrapfig}
\usepackage{dblfloatfix} 
\usepackage[dvipsnames]{xcolor} 
\usepackage{orcidlink} 
\usepackage{enumitem} 
\usepackage{floatrow} 
\usepackage{fancyhdr}
\usepackage{lastpage}

\usepackage[draft]{pdfcomment} 
\usepackage[hyperfirst=false,acronym,sort=none,shortcuts,nopostdot,style=super,nonumberlist,toc,nogroupskip]{glossaries}
\usepackage[dvipsnames]{xcolor} 

\glsdisablehyper 

\newcommand{\accolor}[1]{\textcolor{Sepia}{#1}}

\newacronymstyle{myacro}
{%
  \GlsUseAcrEntryDispStyle{long-short}%
}%
{%
  \GlsUseAcrStyleDefs{long-short}%
}

\setacronymstyle{myacro}

\newcommand*{\tip}[1]{
    \ifglsused{#1}{
      {\pdftooltip{\accolor{\glsentryshort{#1}}}{\glsentrydesc{#1}}}%
    }{%
      \gls{#1}
    }%
}%

\newcommand*{\tips}[1]{
    \ifglsused{#1}{
      {\pdftooltip{\accolor{\glsentryshortpl{#1}}}{\glsentrydesc{#1}}}%
    }{%
      \gls{#1}
    }%
}%

\newacronym{aae}{AAE}{Average Angular Error}
\newacronym{abmof}{ABMOF}{Adaptive Block Matching Optical Flow}
\newacronym{aee}{AEE}{Average Endpoint Error}
\newacronym{aer}{AER}{Address Event Protocol}
\newacronym{aps}{APS}{Active Pixel Sensor}
\newacronym{arms}{ARMS}{(Aperture Robust Multi-Scale flow}
\newacronym{asic}{ASIC}{Application Specific Integrated Circuit}
\newacronym{bhd}{BHD}{Binary Hamming Distance}
\newacronym{bmof}{BMOF}{Block Matching Optical Flow}
\newacronym{bm}{BM}{Block Matching}
\newacronym[description={Multipurpose Block RAM memory module in FPGA}]{bram}{BRAM}{Block RAM}

\newacronym{cis}{CIS}{CMOS Image Sensor}
\newacronym{cnn}{CNN}{Convolutional Neural Network}
\newacronym{cots}{COTS}{Commodity Off-The-Shelf}
\newacronym{cpu}{CPU}{Central Processing Unit}
\newacronym{cs}{CS}{Center-Surround}
\newacronym{csdvs}{CSDVS}{Center Surround Dynamic Vision Sensor}
\newacronym[description={Coarse to Fine hierarchical search strategy used in BMOF}]{ctf}{CTF}{Coarse to Fine}
\newacronym{davis}{DAVIS}{Dynamic and Active pixel Vision Sensor}
\newacronym{dnn}{DNN}{Deep Neural Network}
\newacronym{dram}{DRAM}{Dynamic RAM}
\newacronym[description={Direction Selective model of motion detection in biological vision, usually either Hassenstein-Reichhardt or Barlow-Levick type}]{ds}{DS}{Direction Selective}
\newacronym{dsec}{DSEC}{Driving Stereo Event Camera}
\newacronym{dsp}{DSP}{Digital Signal Processing unit}
\newacronym{dvs}{DVS}{Dynamic Vision Sensor}
\newacronym{edflow}{EDFLOW}{Event-driven Optical Flow}
\newacronym{efast}{EFAST}{Event-Based time surface FAST}
\newacronym{evimo}{EV-IMO}{Event Camera Independently Moving Objects}
\newacronym[longplural={First In First Out memories}]{fifo}{FIFO}{First In First Out memory}
\newacronym{fpga}{FPGA}{Field Programmable Gate Array}
\newacronym{fpn}{FPN}{Fixed Pattern Noise}
\newacronym{fps}{FPS}{Frames Per Second}
\newacronym{fast}{FAST}{Features from Accelerated Segment Test}
\newacronym{gpu}{GPU}{Graphics Processing Unit}
\newacronym{gt}{GT}{Ground Truth}
\newacronym{harms}{hARMS}{hardware Aperture Robust Multiscale flow}
\newacronym{hdl}{HDL}{Hardware Description Language}
\newacronym{hdr}{HDR}{high dynamic range}
\newacronym{hls}{HLS}{High Level Synthesis}
\newacronym{hres}{HRES}{Horizontal Resistor}

\newacronym{imu}{IMU}{Inertial Measurement Unit}
\newacronym{ip}{IP}{Intellectual Property}
\newacronym[description={Corner point in an image}]{kp}{KP}{Keypoint}
\newacronym{lk}{LK}{Lucas-Kanade}
\newacronym[description={DVS OF method that fits a plane to local event cloud}]{lp}{LP}{Local Plane}
\newacronym{mae}{MAE}{Median Angular Error}
\newacronym{mee}{MEE}{Median Endpoint Error}
\newacronym{mree}{MREE}{Median Relative Endpoint Error}
\newacronym{mpeg}{MPEG}{Motion Picture Experts Group}
\newacronym{mvsec}{MVSEC}{Multi Vehicle Stereo Event Camera}
\newacronym{of}{OF}{Optical Flow}
\newacronym{pcb}{PCB}{Printed Circuit Board}
\newacronym{pe}{PE}{Processing Element}
\newacronym{pl}{PL}{Programmable Logic}
\newacronym{prm}{PRM}{Pixel Rendering Module}
\newacronym{ps}{PS}{Processing System}
\newacronym[description={Average Relative Endpoint Error as percentage of flow magnitude}]{aree}{AREE}{Average Relative Endpoint Error}
\newacronym{ram}{RAM}{Random Access Memory}
\newacronym{raft}{RAFT}{: Recurrent All-pairs Field Transforms}
\newacronym{ratp}{RATP}{Recursive Adaptive Temporal Pooling}
\newacronym[description={A Reference Block centered on the event location in the t-d1 slice}]{rb}{RB}{Reference Block}
\newacronym{ree}{REE}{Relative Endpoint Error}
\newacronym{rgc}{RGC}{Retinal Ganglion Cell}
\newacronym[description={Register Transfer Logic intermediate form, consisting of combinational and synchronous register logic cells}]{rtl}{RTL}{Register Transfer Logic}
\newacronym{sad}{SAD}{Sum of Absolute Differences}
\newacronym[description={Search Area for block matching}]{sa}{SA}{Search Area}
\newacronym[description={Surface of Active Event; image of latest event timestamps, same as Timestamp Image}]{sae}{SAE}{Surface of Active Events}
\newacronym{sd}{SD}{Secure Digital}
\newacronym{sits}{SITS}{Speed Invariant Time Surface}
\newacronym[description={Slice-based FAST that uses accumulated event count slices for detecting keypoints}]{sfast}{SFAST}{Slice-based FAST}
\newacronym{slam}{SLAM}{Simultaneous Localization And Mapping}
\newacronym{silc}{SILC}{Speed Invariant Learned Corners}
\newacronym{sm}{SM}{Supplementary Material}
\newacronym[description={System on Chip; FPGA with embedded programmable processor}]{soc}{SoC}{System on Chip}
\newacronym{sram}{SRAM}{Static RAM}
\newacronym{susan}{SUSAN}{Smallest Univalue Segment Assimilating Nucleus}
\newacronym[description={A candidate or best-match Target Block for block matching in the t-d2 slice}]{tb}{TB}{Target Block}
\newacronym[description={Timestamp Image; image of latest event timstamps, same as Surface of Active Events}]{ti}{TI}{Timestamp Image}
\newacronym{timsl}{TS}{time slice}
\newacronym{usb}{USB}{Universal Serial Bus}
\newacronym{vga}{VGA}{Video Graphics Adaptor}
\newacronym[description={Visual Odometry}]{vod}{VOD}{Visual Odometry}
\newacronym{uart}{UART}{Universal Asynchronous Receiver/Transmitter}

\newacronym{dog}{DoG}{Difference of Gaussian}

\usepackage[abbreviations]{foreign} 
\usepackage[style=ieee, backend=biber,maxnames=6]{biblatex} 
\DeclareSourcemap{
  \maps[datatype=bibtex]{
    \map[overwrite]{
      \step[fieldsource=doi, final]
      \step[fieldset=url, null]
      \step[fieldset=eprint, null]
    }  
  }
}

\newcommand{\vp}{V_{\text{p}+}}
\newcommand{\vh}{V_\text{h}}
\newcommand{\vinh}{V_{\text{p}-}}
\newcommand{\ohm}{\Omega}

\AtEveryBibitem{%
  \clearlist{language}
}

\def\BibTeX{{\rm B\kern-.05em{\sc i\kern-.025em b}\kern-.08em
    T\kern-.1667em\lower.7ex\hbox{E}\kern-.125emX}}
    
\fancypagestyle{firststyle}
{
   \fancyhf{}
   \fancyhead[L]{}
   \fancyhead[C]{Submitted to \href{https://2022.ieeeicip.org/}{29th IEEE International Conference on Image Processing (IEEE ICIP 2022)}\\Special session 20.11 `Neuromorphic and perception-based image acquisition and analysis'' }
   \fancyhead[R]{}
  \fancyfoot[C]{\thepage\ of \pageref{LastPage}}
}

\fancypagestyle{otherpages}
{
 
  \fancyhead[R]{}
  \fancyhead[L]{}
  \fancyfoot[C]{\thepage\ of \pageref{LastPage} }
}

\begin{document}
\thispagestyle{firststyle}
\pagestyle{otherpages}

\title{Utility and Feasibility of a Center Surround Event Camera
}
\name{Tobi Delbruck\orcidlink{0000-0001-5479-1141}, Chenghan Li\orcidlink{0000-0003-3523-2787}, Rui Graca\orcidlink{0000-0002-7463-2866}, Brian Mcreynolds\orcidlink{0000-0001-5274-363X}\thanks{C. Li was with UZH-ETH at time of work; is now with inivation.com. Funded by EU project Visualise (600954), Swiss National Science Foundation project SCIDVS (185069), and the USAF Academy Department of Physics, Instructor Faculty Pipeline.}}
\address{Inst. of Neuroinformatics, UZH-ETH ZUrich \\
Zurich, Switzerland. Primary Contact: tobi@ini.uzh.ch.}

\maketitle

\begin{abstract}
Standard dynamic vision sensor (\textbf{DVS}) event cameras output a stream of spatially-independent log-intensity brightness change events so they cannot suppress spatial redundancy. Nearly all biological retinas use an antagonistic center-surround organization. This paper proposes a practical method of implementing a compact, energy-efficient Center Surround DVS (\textbf{CSDVS}) with a surround smoothing network that uses compact polysilicon resistors for lateral resistance. The paper includes behavioral simulation results for the CSDVS (see \href{https://sites.google.com/view/csdvs/home}{sites.google.com/view/csdvs/home}). The CSDVS would significantly reduce events caused by low spatial frequencies, but amplify the informative high frequency spatiotemporal events.



\end{abstract}

\begin{keywords}
retina, pixel, neuromorphic
\end{keywords}

\section{Introduction}

\begin{figure}[tb]
    \centering
    \includegraphics[width=\columnwidth]{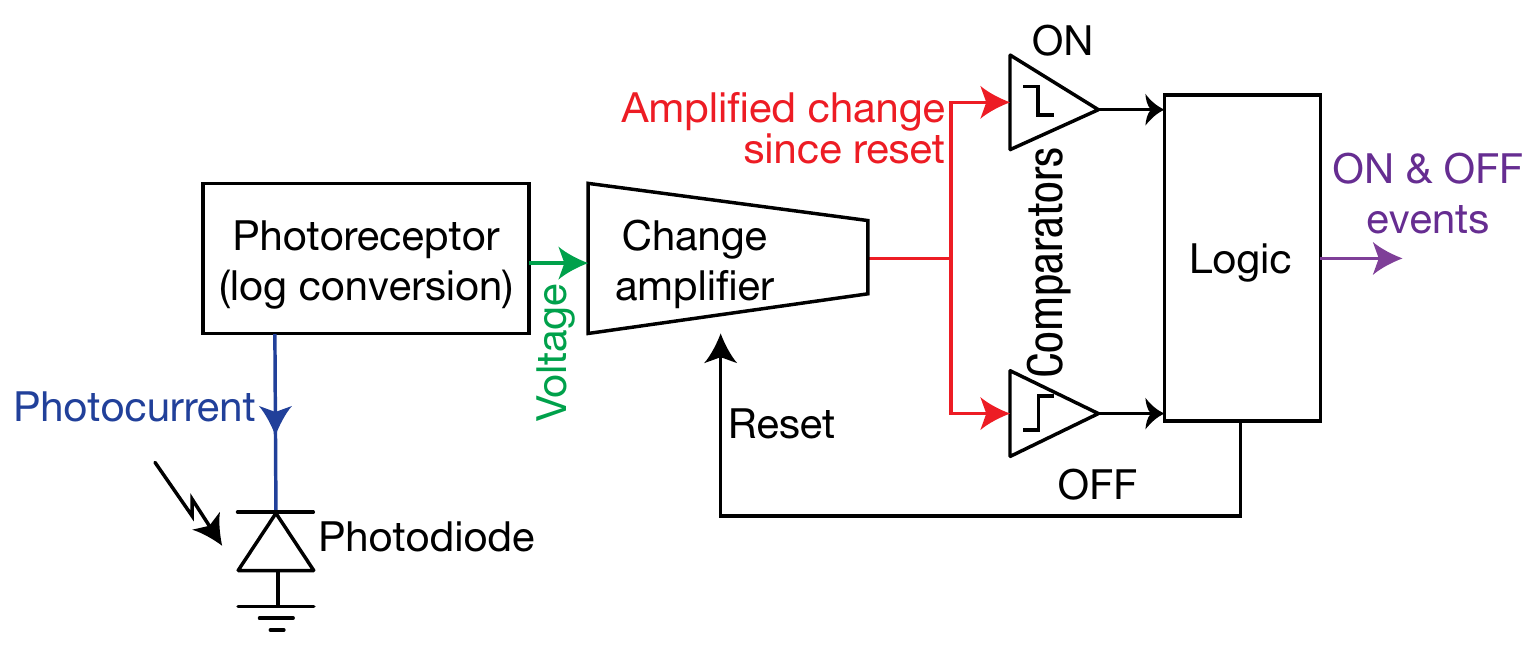}
    \caption{Original \tip{dvs} pixel circuit. 
    Adapted from \cite{Lichtsteiner2008-bk, Li2017-two-streams-phd}.}
    \label{fig:dvs-circuit}
\end{figure}

  \tip{dvs} event cameras~\cite{Lichtsteiner2008-bk} based on the pixel architecture illustrated in Fig.~\ref{fig:dvs-circuit} output a stream of pixel-level brightness change events that are proving useful for quick, low power, high dynamic range vision applications~\cite{Gallego2020-vs}. 

A shortcoming of the \tip{dvs} is that it responds to all local brightness changes, generating output from any moving image gradient---even low frequency and relatively uninformative gradients---and from brightness changes caused by fluctuating illumination. Many artificial lighting systems (sodium, LED, and fluorescent) flicker at some frequency. These lighting fluctuations can cause a storm of \tip{dvs} events that are largely uninformative, requiring either increasing the \tip{dvs} event threshold, decreasing the photoreceptor bandwidth, or both, limiting the ability to transmit informative events about scene reflectance changes.

This problem can be solved by imitating a key feature of mammalian retinas. They have a lateral ``surround'' network consisting of a 2D mesh of horizontal cells that are connected by conductive gap junction synapses~\cite{Dowling2012-retina-revised-edition}. The surround averages local photoreceptor activity over space and time. This photoreceptor and surround are combined antagonistically with a \tip{cs} arrangement.
 This way, the surround can cancel the output activity from uniform regions. So far, \tip{dvs} cameras have not included such \tip{cs} architecture because the functional advantages were not evident and it was not known how to implement it in a compact and precise form.

This paper proposes a compact and energy-efficient \tip{csdvs} design.
The key contribution of this paper is a new compact surround design. Instead of using bulky and imprecise transistors for lateral surround resistors (like past designs), the surround consists of fixed lateral polysilicon resistors combined with a controllable transverse transconductance. The response of the resulting surround will be effectively instantaneous, and its size can be controlled over a wide range.
The \tip{csdvs} pixel would increase the circuit area by about 20\%, but would significantly decrease low spatial frequency output, particularly in response to fluctuating illumination.

\begin{figure*}
\floatbox[{\capbeside\thisfloatsetup{capbesideposition={left,center},capbesidewidth=7cm}}]{figure}[\FBwidth]
{\caption{\glsreset{csdvs}\tip{csdvs} pixel circuit that produces spatially filtered brightness change events. It is tiled to connect the lateral resistors to neighboring pixels.}\label{fig:csdvs-circuit}}
{\includegraphics[width=.55\textwidth]{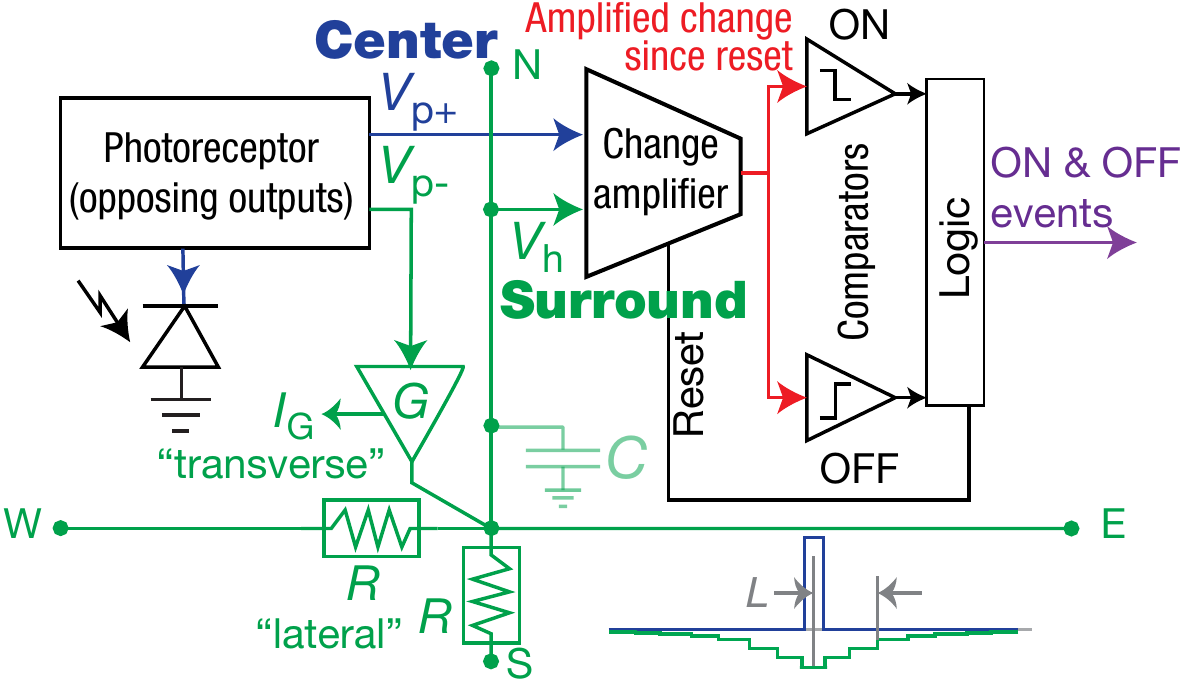}}
\end{figure*}

Fig.~\ref{fig:csdvs-circuit} illustrates the \tip{csdvs} pixel circuit. It uses a resistive network driven by a transverse conductance $G$ from the inverted photoreceptor signal $\vinh$ to represent the antagonistic horizontal cell surround signal $\vh$.  The difference $\vp-\vh$ suppresses events from groups of pixels with similar photoreceptor output. The operating principle and circuit are described in Secs.~\ref{sec:principle} and \ref{sec:pixel}.


\section{Related Work}\label{sec:related}

Neuromorphic engineers have long dreamed of implementing an electronic model of the biological retina. Fukushima's discrete electronics model from the 1970s used a resistor mesh to model the horizontal cell network~\cite{Fukushima1970-electronic-retina}. The earliest integrated silicon retinas of Mahowald and Mead~\cite{Mead1988-retina-neural-networks,Mead1989-avlsi,Mead1989-adaptive-retina-chapter,Mahowald1991-spie-retina-adaptive,Mahowald1992-phd-thesis} featured a horizontal spreading network constructed from transistors using Mead's \tip{hres} circuit~\cite[Ch. 7]{Mead1989-avlsi}. 
Numerous other imaginative silicon vision sensors featured transistor-based spatial and spatiotemporal filtering at the focal plane~\cite{Boahen1997-phd-thesis,Liu2000-fly-global-processing,Harrison2000-reichhardt,Delbruck2004-friend,Zaghloul2006-silicon-retina-spiking-contrast-gain-control,Costas-Santos2007-spatial-constrast-calib, Lenero-Bardallo2010-signed-spatial-calibration}.
These devices had complex pixels (which reduced resolution) and lots of transistor mismatch, which produced excessive salt and pepper \tip{fpn} in the output.
In-pixel digital calibration circuits were bulky and limited to compensating a single current mirror~\cite{Costas-Santos2007-spatial-constrast-calib, Lenero-Bardallo2010-signed-spatial-calibration}. At the same time, the computer vision community was being treated to clean megapixel \tip{cis} cameras, Moore's law, and the Internet, so silicon retina development largely stalled out, despite some persistent and beautiful work from the Yagi lab in Osaka~\cite{Yagi1999-parallel-analog,Kameda2003-analog-retina} and Ruedi at CSEM~\cite{Ruedi2003-cg,Ruedi2009-uh}.


Still, \tip{cis} cameras had---and still have---nagging problems of limited sample rate, limited dynamic range, redundant output, motion blur, and power-latency trade off~\cite{Liu2019-eds-ieee-sig-proc}. These problems kept interest alive. Since the mid-2000s, \tips{dvs} and subsequent event cameras using the active logarithmic photoreceptor and switched capacitor change detector from \cite{Lichtsteiner2008-bk} reduced \tip{fpn} and improved overall
performance, opening up the field for event camera applications~\cite{Gallego2020-vs}. Some event cameras even include activity-driven~\cite{Posch2011-wm} or sampled~\cite{Brandli2014-davis} intensity values. Resolution and readout bandwidth have increased dramatically, which is great for applications like self-driving cars that must see small objects far away~\cite{Christensen2022-neuromorphic-roadmap}.     

However, biological \tip{rgc} spikes are the result of vastly sophisticated computations relevant for survival, and it seems with the drive for more and smaller pixels, event camera design is diverging from biology.  With this rich historical background---and with industry occupied with the megapixel \tip{dvs} race~\cite{Suh2020-samsung-dvs-1280x960,Finateu2020-prophesee-isscc,Christensen2022-neuromorphic-roadmap}---it is a good opportunity to revisit the idea of implementing at least a simple antagonistic center surround.
The main questions we answer are:
\begin{enumerate}[noitemsep]
    \item Would a center-surround event camera be useful?
    \item Is it possible to design a compact and precise pixel?
\end{enumerate}

\noindent Preliminary work was reported in Li's PhD thesis~\cite[Chapter 5.1]{Li2017-two-streams-phd}. This paper extends that report by a better understanding of pixel dynamics and behavioral simulation.

\section{Functional principle}\label{sec:principle}

The \tip{cs} computation (illustrated by the inset in Fig.~\ref{fig:csdvs-circuit}) takes the difference between 
the photoreceptor output signal $\vp$ and the average $\vh$ of the surrounding neighbors over a 
 space constant $L$, where $L$ is the distance from the origin of an input to the node where the surround signal has decayed to  $1/e$ of the signal at the origin node.
This \tip{cs} computation is a spatial highpass filter; spatial frequencies lower than $\sim1/L$ pixels are filtered out. 

Past \tip{cs} silicon retinas have mainly output the static difference between the $\vp$ center and the $\vh$ surround, modeling
sustained \tips{rgc}~\cite{Dowling2012-retina-revised-edition}. 
The \tip{csdvs} pixel models a transient retinal pathway such as is found in most peripheral \tips{rgc}:
Its output consists of asynchronous changes in $\vp-\vh$ that exceed the ON and OFF threshold $\theta$:
\begin{equation} \label{eq:threshold}
    |\Delta (\vp-\vh)|>\theta.
\end{equation}
After each event, the value of $\vp-\vh$ is memorized by the change detector.


\section{Functional Utility} \label{sec:simulations}

\begin{figure}
    \centering
    \includegraphics[width=.8\columnwidth]{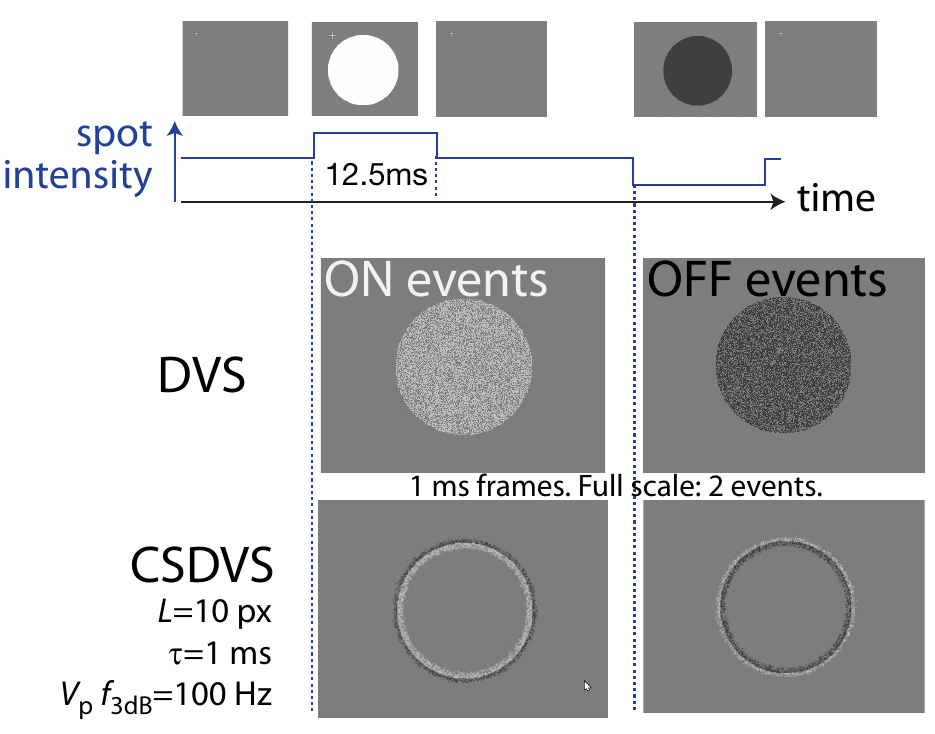}
    \caption{Comparison of simulated normal \tip{dvs} and \tip{csdvs} response to a flashing spot.}
    \label{fig:spots}
\end{figure}

\begin{figure}[tb]
    \centering
    \includegraphics[width=\columnwidth]{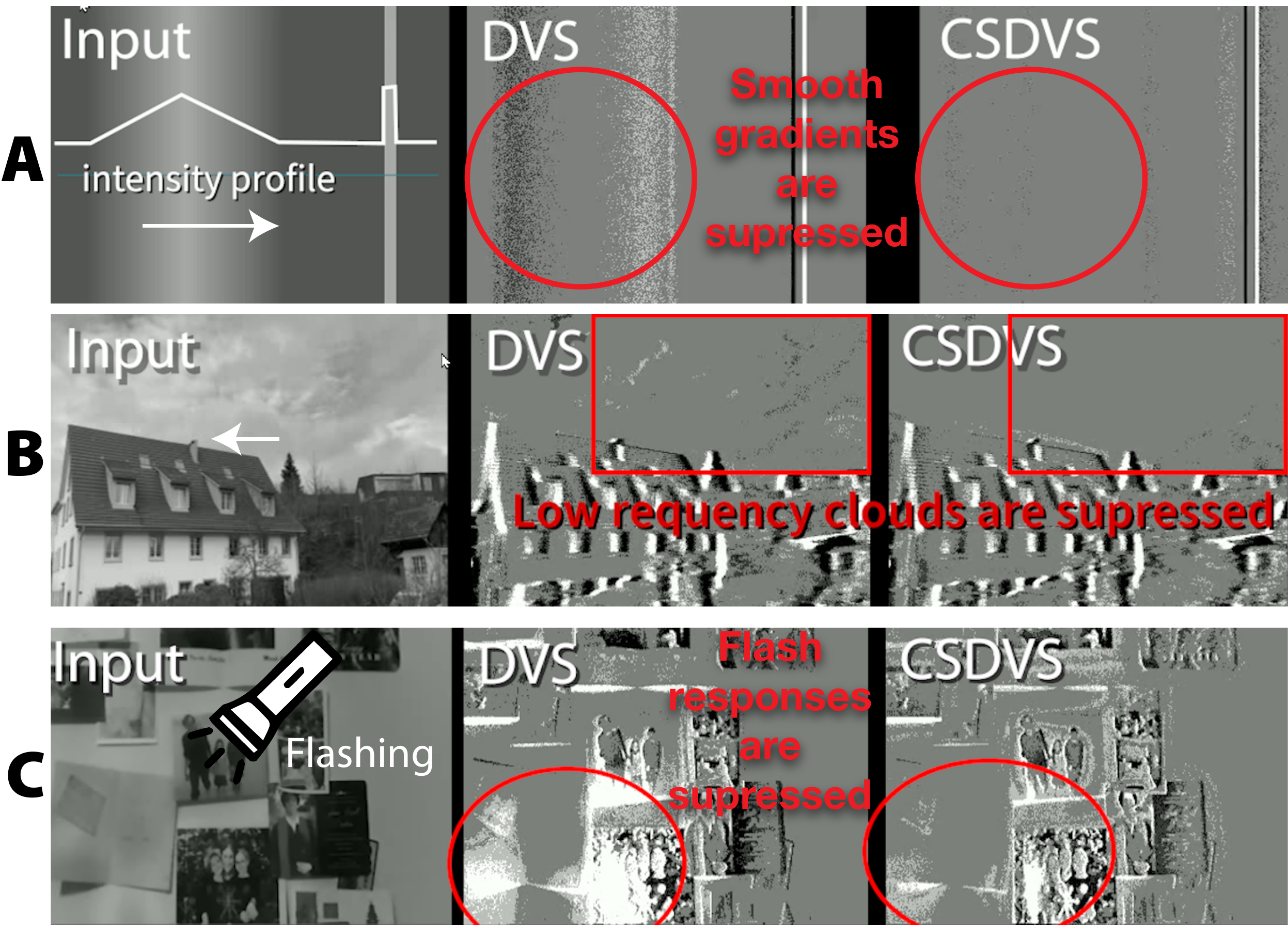}
    \caption{Comparison of normal \tip{dvs} and \tip{csdvs} simulated responses with highlighted differences. \textbf{A:}  Flashing spot. \textbf{B:} Moving gradient. \textbf{C:} Panned outdoor cloudy scene. \textbf{D:} Flashing illumination.}
    \label{fig:v2e-samples}
\end{figure}

To model \tip{csdvs}, we branched\footnote{\href{https://github.com/SensorsINI/v2e/tree/CS-DVS}{CS-DVS branch of v2e on github}} our video to events camera simulator \textsl{v2e}~\cite{Hu2021-v2e-ieee} to include an optional antagonistic surround network. 
A user can specify the space constant $L$ and time constant $\tau=C/G$, where is $C$ is the capacitance on each $\vh$ node.
\textsl{v2e} simulates the \tip{csdvs} event output in response to a standard frame video input. This section shows the results of these simulations. 

Source videos were either generated synthetically or captured by a digital video camera. 
For the \tip{csdvs} simulation, we used either $L=10\text{ px}$ or $L=30\text{ px}$ and $\tau=2\text{ms}$\footnote{Sec.~\ref{sec:pixel} shows that $\tau$ should be much smaller, but it makes the Euler stepping of \eqref{eq:ode} extremely slow.}. We assumed $\theta=0.2\pm0.02$ natural log units of intensity and used a photoreceptor $f_\text{3dB}$ cutoff frequency of 100\,Hz to model a \tip{dvs} under low illumination.

Readers are invited to view videos of the following \tip{csdvs} vs.\ \tip{dvs} simulation results\footnote{See \href{https://sites.google.com/view/csdvs/home}{sites.google.com/view/csdvs/home}}.

Fig.~\ref{fig:spots} shows flashing spot responses (see \textit{spots} video for the complete sequence including moving spots). 
The spot flashes from gray to bright and then back to gray, then to dark, and back to gray. 
The spot contrast is 1.5. 
The normal \tip{dvs} makes ON and OFF events over the entire spot and none outside it; however, the \tip{csdvs} produces events only at the edges of the spot. At the center of the spot, the surround responds nearly identically to the photoreceptor, suppressing events from this uniform area. For pixels just inside and just outside of the spot, the $\vh$ response is opposite to the $\vp$ response, resulting in more events (signal amplification) than from the \tip{dvs}.
In total, the \tip{dvs} simulation produced 302k events, and the \tip{csdvs} only 124k events, a reduction of about 60\%.

Fig.~\ref{fig:v2e-samples}A (\textit{gradients} video) shows the response to two synthetic moving bumps of equal contrast. The left bump is gradual and the right bump is sharp. \tip{csdvs} filters out the low-frequency gradual edge but slightly amplifies the sharp edge.

Fig.~\ref{fig:v2e-samples}B (\textit{cloudy-sky} video) compares response to a natural outdoor scene with a partially clouded sky. The source video was captured by a smartphone during a rapid 1s pan over the scene. This scene consists mostly of high frequency content, so the \tip{dvs} and \tip{csdvs} event rates are similar (4.8\,MHz vs.\ 3.6\,MHz), but the low frequency clouds are erased while informative, high frequency information is retained.

Finally, Fig.~\ref{fig:v2e-samples}C (\textit{flicker} video) compares the response to a scene with global flicker caused by flashing lighting.
Where the scene has high frequency spatial contrast, the difference between the center and surround produces activity. However, in uniform areas, the difference between \tip{dvs} and \tip{csdvs} response is dramatic: \tip{dvs} produces massive activity from the flashing but \tip{csdvs} suppresses it.

\section{Pixel Design}\label{sec:pixel}

Mead~\cite[Chapter 7]{Mead1989-avlsi} showed that a one-dimensional discrete resistive network has a
space constant length $L$ that has the relation \eqref{eq:rgspaceconstant} with the transverse conductance $G$
and the lateral resistance $R$:
\begin{equation}
    \label{eq:rgspaceconstant}
    L=\frac{1}{\sqrt{RG}}.
\end{equation}

For a 2D resistive mesh, 
\eqref{eq:rgspaceconstant} still describes the response to an edge~\cite[Ch.~7,App.~C]{Mead1989-avlsi}. 
Feinstein's analysis showed for a 2D mesh that \eqref{eq:rgspaceconstant} still approximately holds for $L\gg 1$~\cite{Feinstein1988-au}. 

\subsection{Time domain} \label{subsec:freqdomain}
Fig.~\ref{fig:csdvs-circuit} outlines a potential \tip{csdvs} circuit.
Each horizontal cell surround node $\vh$ is driven by the inverted photoreceptor output $\vinh$ through transconductor $G$. It has dynamics determined by \eqref{eq:ode}:
\begin{equation} \label{eq:ode}
    C \frac{d\vh}{dt} =G (\vp-\vh)-\frac{1}{R}\sum\limits_{j=\text{NSEW}}(\vh-V_j),
\end{equation}
\noindent where $C$ is the capacitance, and $\text{NSEW}$ means the 4 nearest neighboring $\vh$ nodes.

\subsection{Design of surround}
\label{subsec:csdesign}

Previous \tip{cs} retina designs used some type of CMOS transistor surround, which allowed control of surround space and time constant independently. This choice required many transistors and the transistor mismatch caused a lot of \tip{fpn}.
But \eqref{eq:rgspaceconstant} shows that the larger the space constant $L$,
the smaller the transconductance $G$ needs to be, which means it might be possible to use normal (fixed) resistors for $R$ and to implement $G$ with transistors. That would still allow control of $L$ but the resistors would be more compact and better matched.
Therefore, we reconsider using polysilicon resistors for the lateral resistor $R$, rather than the bulky \tip{hres} circuit. 
The computations in this section are order-of-magnitude since we do not yet have detailed  circuit implementations.

Unsalicided polysilicon\footnote{Salicide is a highly conductive alloy of metal and silicon that normally decreases the polysilicon gate conductor resistance. If the metalization is blocked, the conductance is controlled by the polysilicon doping.} is a standard resistor device offered in many processes. For example, in a 180nm process we have used for \tip{dvs} chips, 
the unsalicided polysilicon
 has a sheet resistance of 2k$\ohm/\square$
 and a 3$\sigma$ mismatch of 5\% for a
2um wide 100um long resistor. 
If we use 1um wide and 5um long unsalicided
polysilicon as one resistor (to fit along one edge of a $5\times 5\text{um}^2$ pixel), its resistance is 10k$\ohm$, with a likely $3\sigma$ matching to 10\%. 

Now \eqref{eq:rgspaceconstant} means that $G\sim 1/(R L^2)$. 
The transconductor $G$ could be implemented with a 5-transistor transconductance amplifier or even a 2-transistor source follower, since the DC level is ignored by the \tip{dvs} change detection.
Since subthreshold transconductance $G$ is related to bias current $I_\text{G}$ by 
 $G\sim I_\text{G}/U_\text{T}$, then
 \begin{equation}
     \label{eq:ivsl}
     I_\text{G} \sim \frac{U_\text{T}}{R L^2}\text{ and }L\sim\sqrt{\frac{U_\text{T}}{RI_\text{G}}},
 \end{equation}
 where $U_\text{T}$ is the thermal voltage (25\,mV  at room temperature).
 Thus, for $L=10\text{ px}$,
 with $R=10\,\text{k}{\ohm}$,
 $I_\text{G} \sim 10\text{ nA}$, which is a small current.
For a megapixel array, the total $G$ bias current would only be 10mA, a fraction of the total supply current. 

\begin{figure}
    \centering
    \includegraphics[width=.8\columnwidth]{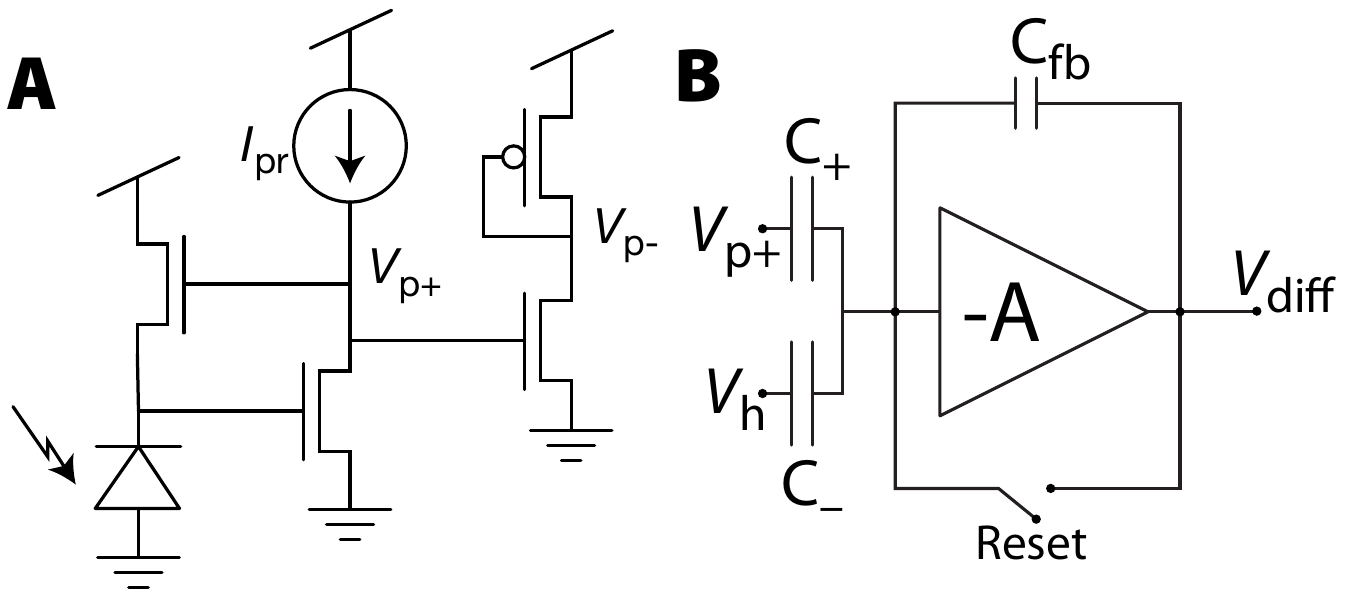}
    \caption{\textbf{A:} \tip{csdvs} photoreceptor circuit that produces opposing output voltages $\vp$ and $\vinh$. \textbf{B:} Summing switched capacitor change detector amplifier. Adapted from \cite{Li2017-two-streams-phd}.}
    \label{fig:circuits}
\end{figure}

The surround temporally low-pass filters the photoreceptor through $\tau=C/G$. 
However, because the lateral resistance $R$ is only a few k$\ohm$, 
$G\sim 1/(R L^2)$ is still large. 
$C$~mainly consists of the summing amplifier input capacitance $C_{-}$ (Fig.~\ref{fig:circuits}B).
Overestimating $C=1\text{pF}$
and $R=100\text{k}\ohm$
and taking $L=10$, 
then 
\begin{equation} \label{eq:tau}
    \tau\sim C/G= RCL^2=~10^5\ohm \times 1^{-12}\text{F}\times 10^2=10\text{us},
\end{equation}
\noindent corresponding to a cutoff frequency
$f_\text{3dB}=1/(2\pi\tau)\approx 10\text{kHz}$, 
which is about 10X faster than the maximum photoreceptor bandwidth.
The surround would effectively respond instantaneously to the photoreceptor input, allowing the cancellation of redundant events caused by flickering illumination.

To detect changes in the difference $\vp-\vh$ according to \eqref{eq:threshold}, the surround must be subtracted from the photoreceptor. Fig.~\ref{fig:circuits}A shows a photoreceptor circuit that produces opposing outputs $\vp$ and $\vinh$; the $\vinh$ drives the input to the surround. 
Fig.~\ref{fig:circuits}B shows a switched capacitor circuit that sums the positive photoreceptor and negative surround voltages to detect significant events. The relative sizes of $C_+$ and $C_-$ are adjusted by circuit simulation to compensate for the small $\vp$ and $\vinh$ gain differences. The large $L$ reduces the effect of expected large $G$ mismatch.

\section{Conclusions}\label{sec:conclusion}


The proposed \tip{csdvs} design would provide a surround with a controllable size. The surround would effectively be instantaneous.
\tip{csdvs} would amplify high spatial frequencies and significantly reduce \tip{dvs} activity in uniform and smoothly varying areas of the scene. It would have a dramatic effect on reducing activity in uniform areas of the scene caused by time-varying lighting.

The proposed surround would add 4-8  transistors (depending on the transconductor design) and 2 narrow stripes of polysilicon resistor. 
 The \tip{csdvs} pixel would use about 10 fewer large analog transistors than a design using the \tip{hres}. Hence, the \tip{csdvs} design
would be feasible with a modest increase in pixel complexity. 
Combined with the switched capacitor \tip{dvs} change detection, we expect much less \tip{fpn} than in past \tip{cs} silicon retinas.

Although \tip{csdvs} (like \tip{dvs}) remains a gross simplification of the vastly more sophisticated biological retina, it may form the next useful abstraction in the delicate dance between introducing features and increasing pixel complexity. It may be beneficial to dynamically aggregate photoreceptors for the center and to dynamically modulate both center and surround radii as is seen in biology~\cite{Dowling2012-retina-revised-edition}, and to exploit $\vh$ feedback in some manner like~\cite{Mahowald1992-phd-thesis, Zaghloul2006-silicon-retina-spiking-contrast-gain-control}.

\textit{Acknowledgments --} We thank SC Liu for useful comments.

\printbibliography

\end{document}